\documentclass[3p,times]{elsarticle}

\usepackage[bookmarks=false]{hyperref}
    \hypersetup{colorlinks,
      linkcolor=blue,
      citecolor=blue,
      urlcolor=blue}
\usepackage{orcidlink}

\newif\ifanonymous
\anonymousfalse

\journal{arXiv}

\usepackage{amssymb}
\usepackage{amsmath}
\usepackage{amsthm}
\usepackage{booktabs}
\usepackage[figuresright]{rotating}

\biboptions{numbers,sort&compress}

\newtheorem{proposition}{Proposition}
\newtheorem{definition}{Definition}

\newcommand{\proc}{EGP}
\newcommand{\elide}{ELIDE}
\newcommand{\Vstop}{\mathcal{V}_{\mathrm{stop}}}
\newcommand{\pent}{p_{\mathrm{ent}}}
\newcommand{\smax}{s_{\max}}
\newcommand{\TRR}{\mathrm{TRR}}

\begin{document}
\begin{frontmatter}

\title{Beyond Truncation: When Can LLMs Safely Omit Generated Text?}

\ifanonymous
  \author{Anonymous Author(s)}
  \address{Anonymous Institution}
\else
  \author[a]{Mahmoud Alyosify\,\orcidlink{0009-0005-1336-2726}\corref{cor1}}
  \author[a]{Mirna Imbabi\,\orcidlink{0009-0000-6699-3170}}
  \author[a]{Eman Elkhamisi\,\orcidlink{0009-0002-6077-3593}}
  \author[a]{Rahatara Ferdousi\,\orcidlink{0000-0003-1143-2370}}
  \address{School of Computing, Queen's University, Kingston, Ontario, Canada \\
    \texttt{\{mahmoud.alyosify, mirna.imbabi, eman.elkhamisi, rahatara.ferdousi\}@queensu.ca}}
\fi

\begin{abstract}
Instruction-tuned large language models (LLMs) answer at greater length than a request needs. We ask whether a training-free controller inside the decoding loop can remove that surplus while preserving content, judged against the simplest possible baseline: truncating the model's own answer to the same length. An entropy-gated logits processor removes 25.5--34.9\% of tokens on MT-Bench development questions and 19.9--34.2\% on a sealed test split within pre-registered BERTScore non-inferiority margins, yet we prove and confirm by audit that its output is always a prefix of the unmodified answer: the saving is truncation. A survey of 23 training-free mechanism families finds none that selects content better than truncation without oracle labels. We therefore introduce Entailment-Licensed In-Decoding Elision (\elide{}), which, at each sentence boundary, withholds a span from the reader only when a natural-language inference model judges it entailed by the text already shown, while leaving the model's generation unchanged. \elide{} beats matched-length truncation on three model families, but reading 936 of its elisions on 670 public prompts shows that 59\% harm the answer, a failure that no automatic measure detects. Scoping the licence to its discourse unit reduces the harmful share to about one elision in ten on two fresh tests.\end{abstract}


\ifanonymous\else
\cortext[cor1]{Corresponding author.}
\fi
\end{frontmatter}

\ifanonymous\else
\email{mahmoud.alyosify@queensu.ca}
\fi

\section{Introduction}
\label{sec:intro}

Each generated token costs a full forward pass, bound by memory bandwidth at small batch sizes \cite{pope2023efficiently}, so the time, energy and price of an answer grow roughly linearly with its length (about 1.9\,J per token for Mistral-7B on our A40 GPU), and length also costs the reader's attention. Yet instruction-tuned models are systematically pushed towards length: reward models correlate strongly with output length \cite{singhal2023long}, LLM judges prefer longer answers \cite{saito2023verbosity, zheng2023judging}, and leaderboards have had to introduce length-controlled scores \cite{dubois2024length}.

Optional length is not redundant length: the instruction ``Be concise'' shortens answers of three open models by 36--50\% in our data, but by writing a different, smaller answer, not by deleting restatements. We frame the task precisely: given a frozen instruction-tuned LLM, reduce the tokens it produces, or the tokens shown to the reader, by acting inside the decoding loop without changing the model's weights, such that the result retains the content of the unmodified answer \emph{and retains more of it than the unmodified answer truncated to the same number of tokens}. The last clause is the crux. Any method that ends answers early reduces tokens, and its quality is then exactly the quality of truncation; only by beating that baseline can a method claim to have reduced \emph{verbosity} rather than \emph{length}.

Our contributions are: (i) an entropy-gated logits processor (\proc{}) that clears its pre-registered quality margins, with a proof and a decision-level audit showing that it is a stop controller whose output is always a prefix of the baseline; (ii) the \emph{truncation frontier} as the evaluation bar for any method that shortens text, with a survey of 23 training-free mechanism families against it; (iii) \elide{}, a span-level omission method with an entailment licence borrowed from the linguistics of ellipsis and a trajectory-invariance guarantee, which beats matched-length truncation on three model families; and (iv) a pre-registered reading audit at scale showing that most of its elisions harm the answer although automatic measures approve them, and a discourse-scoped version that lowers the harm roughly sixfold. Every stage was pre-registered and its configurations frozen before evaluation.

\section{Related work}
\label{sec:related}

Length control has mostly been learned, by conditioning on a target length \cite{kikuchi2016controlling} or by fine-tuning models to compress chain-of-thought traces \cite{xia2025tokenskip}; we study the training-free setting. Decode-time interventions such as repetition penalties \cite{keskar2019ctrl}, contrastive decoding \cite{li2023contrastive} and activation steering change which tokens are produced, while quantisation and speculative decoding \cite{leviathan2023speculative} make each token cheaper without changing how many there are, and are complementary to verbosity control. NLI models \cite{williams2018broad} are widely used to check factual consistency, but rely on lexical-overlap heuristics \cite{mccoy2019right}. In linguistics, an ellipsis is licensed only when the omitted material is recoverable from an antecedent \cite{hankamer1976deep}; we operationalise recoverability as entailment by the text already shown.

\section{The truncation frontier}
\label{sec:frontier}

\begin{definition}[Truncation baseline and frontier]
For a method $A$ that produces $n_A^{(i)}$ tokens on prompt $i$, its \emph{truncation baseline} is the first $n_A^{(i)}$ tokens of the unmodified answer $B^{(i)}$. $A$ \emph{beats the frontier} on prompt $i$ if a content measure, computed against $B^{(i)}$, scores $A$ above its truncation baseline. The \emph{truncation frontier} is the curve of content retained against token reduction traced by truncating $B$ at every fraction.
\end{definition}

A method may claim to reduce verbosity only if it beats its truncation baseline, paired per prompt, at exactly matched length. For span-level methods we add \emph{length-only deletion} (whole sentences removed longest first) and \emph{random deletion} at the same length. The comparison costs nothing and, as Section~\ref{sec:res-proc} shows, overturns results that pass conventional quality margins.

\section{Methods}
\label{sec:methods}

\subsection{An entropy-gated logits processor}
\label{sec:proc}

\proc{} is a single logits processor placed between the language-model head and token selection. At decoding step $t$, with logits $z_t$ over the vocabulary, it returns
\begin{equation}
  \tilde{z}_t(v) = z_t(v) - g(H_t)\,\bigl[\, w_L\,L_t(v) + w_R\,R_t(v) + w_S\,S_t(v) \,\bigr],
  \label{eq:processor}
\end{equation}
where the gate $g(H_t)=\mathrm{clip}\bigl((H_t-H_{\mathrm{low}})/(H_{\mathrm{high}}-H_{\mathrm{low}}),0,1\bigr)$ opens with the entropy $H_t$ of the model's next-token distribution. The \emph{length term} is non-zero only at stop tokens: $L_t(v)=-\ell(t)\,m_t$ for $v\in\Vstop$, with a ramp $\ell(t)=\mathrm{clip}\bigl((t-t_{\min})/(t_{\max}-t_{\min}),0,1\bigr)^{\gamma}$ and the margin $m_t=\max_v z_t(v)-\max_{s\in\Vstop} z_t(s)\ge 0$ that the best stop token must close. The \emph{repetition term} $R_t(v)\in\{0,1\}$ marks candidates completing an already generated trigram, and the \emph{semantic term} $S_t(v)$ is a thresholded cosine between a candidate's unembedding row and those of the generated tokens. Calibration showed that the semantic term changed the chosen token on only 12 of 24,021 steps at any weight and that the repetition term lengthened answers, so the frozen operating point uses the length term alone ($w_L=1.1$, $\gamma=1.5$, $(H_{\mathrm{low}},H_{\mathrm{high}})=(0.05,0.6)$ nats, $w_R=w_S=0$), with $t_{\min}$ and $t_{\max}$ set from each model's baseline length percentiles. Its structure determines what it can do:

\begin{proposition}[Frozen-point equivalence]
\label{prop:frozen}
If $w_R=w_S=0$, then $\tilde z_t(v)=z_t(v)$ for every $v\notin\Vstop$. Greedy decoding under the processor therefore selects, at every step, either the token the unmodified model selects or a stop token, and its output is a prefix of the unmodified output followed by a stop token. Moreover, the best stop token overtakes the unmodified argmax at step $t$ if and only if $g(H_t)\,w_L\,\ell(t)>1$.
\end{proposition}
\begin{proof}[Proof sketch]
Non-stop logits are untouched, so the processed argmax is either the unmodified argmax or a stop token whose logit was raised; induction over steps gives the prefix property. The best stop logit becomes $\max_s z_t(s)+g\,w_L\,\ell\,m_t$, which exceeds $\max_v z_t(v)=\max_s z_t(s)+m_t$ exactly when $g\,w_L\,\ell>1$; the margin cancels.
\end{proof}

The flip condition contains nothing about what the answer says: at $w_L=1.1$ the processor ends an answer at the first step after token 377--396 (by model) at which entropy exceeds about 0.55 nats.

\subsection{Entailment-Licensed In-Decoding Elision (\elide{})}
\label{sec:elide}

\begin{figure}[!htb]
\centering
\includegraphics[width=0.66\textwidth]{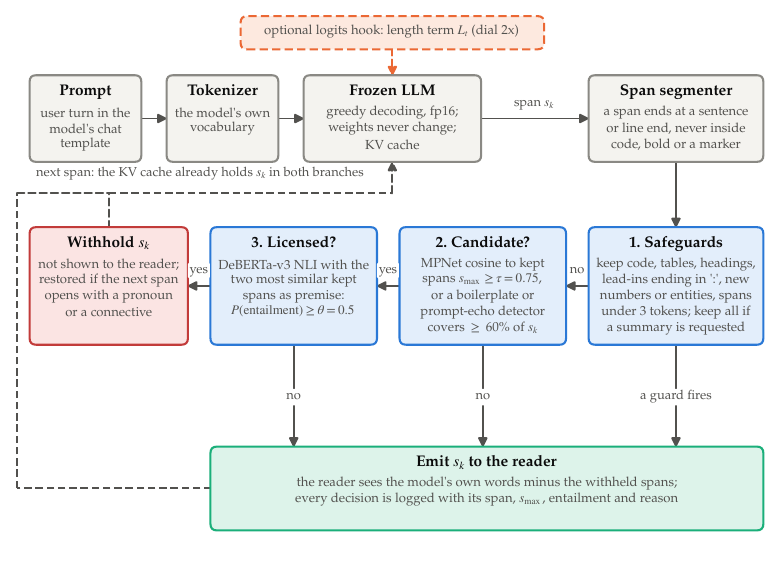}
\caption{\elide{}. Each completed span passes the safeguards, the cosine pre-filter and the NLI licence; a licensed span is withheld from the reader but stays in the KV cache, so the model's generation is unchanged. The processor's length term can optionally be attached as a logits hook.}
\label{fig:arch}
\end{figure}

The survey of Section~\ref{sec:res-survey} yields three design constraints: the model cannot detect its own redundancy, so detection must come from outside instruments; every mechanism that edits non-stop tokens either reroutes or drifts, so the method must never touch a logit; and redundancy is a relation between a sentence and what has already been said, so the decision belongs at sentence boundaries. \elide{} is therefore an \emph{omission} actuator (Fig.~\ref{fig:arch}). The answer is cut into spans as it is generated, a span ending at a sentence boundary (list markers, decimals and abbreviations excluded), at a line break, or after 48 tokens. Each completed span is either shown to the reader or withheld; in both cases it remains in the model's context.

\paragraph{Candidates} Let $K$ be the spans already shown and $e(\cdot)$ the normalised all-mpnet-base-v2 embedding \cite{reimers2019sentence} of a span with its markdown removed. A new span $s_k$ becomes a candidate when $\smax(s_k)=\max_{u\in K}\cos(e(s_k),e(u))\ge\tau=0.75$. The encoder separates restating from novel spans well (held-out AUROC 0.988), but it also flags sentences that share a template and differ in content, so it only decides whether the licence is consulted.

\paragraph{The licence} Let $u_1u_2$ be the two kept spans most similar to $s_k$, concatenated in their original order. The candidate is \emph{licensed} when $\pent(\textit{entailment}\mid\text{premise}=u_1u_2,\ \text{hypothesis}=s_k)\ge\theta=0.5$, where $\pent$ is DeBERTa-v3-base fine-tuned on MultiNLI, FEVER and ANLI \cite{he2023debertav3}. Markdown is stripped from premise and hypothesis: on a control, the entailment probability of a duplicated list item rose from 0.009 with list markers to 0.995 without them.

\paragraph{Safeguards} A licensed candidate is still kept if it is code, a table row, a heading, a list marker or a lead-in ending in a colon; if the user asked for a summary or verbatim text; if it is under three tokens or names a number or entity absent from the kept text; or after 3 consecutive or 32 total omissions. An omitted span is restored if the next span opens with a pronoun or connective, and its line breaks are carried forward so that lists keep their shape.

\begin{proposition}[Trajectory invariance]
\label{prop:traj}
\elide{} generates exactly the greedy answer $y$ of the unmodified model and shows $y$ with the withheld spans removed.
\end{proposition}
\begin{proof}[Proof sketch]
Whether a span is shown never enters the model's context: in both branches the KV cache advances through the span's tokens, so each span is decoded from the same state as in the unmodified run, and by induction the generated sequence is $y$.
\end{proof}
Consequently \elide{} cannot reroute: its effect is exactly the set of withheld spans, each logged with the cosine and entailment values that justified it. It saves \emph{shown} tokens, i.e.\ reading and downstream token-billed processing, rather than generation.

\paragraph{Version 2: a scoped licence} Reading the first version's (v1) elisions at scale (Section~\ref{sec:res-scale}) motivated five restrictions, each of which only shrinks the set of withheld spans: (1)~\emph{item scope}: a span inside a list item may be licensed only by kept text of the same item, the locality condition of ellipsis at the granularity of the answer's markup; (2)~no prompt-echo rule; (3)~a six-word minimum; (4)~a genre guard for verse, scripts, dialogue and structured formats; and (5)~stricter segmentation and qualifier guards. Thresholds are unchanged.

\section{Experimental setup}
\label{sec:setup}

\paragraph{Models and data} Three instruction-tuned models were pinned by revision: Mistral-7B-Instruct-v0.2 \cite{jiang2023mistral}, Llama-3.1-8B-Instruct \cite{dubey2024llama3} and Qwen2.5-3B-Instruct \cite{qwen2024qwen25}; OLMo-2-7B-Instruct \cite{olmo2024olmo2} and Granite-3.1-8B-Instruct \cite{granite2024granite} test transfer. Decoding is greedy in fp16 at batch size one on one NVIDIA A40. The processor is evaluated on the 80 MT-Bench questions \cite{zheng2023judging}, split by a frozen hash into 16 development questions and 64 test questions that were sealed until every operating point was frozen. \elide{} was selected on 8 hand-written multi-part prompts, verified on 24 held-out ones, scaled to a 670-prompt development frame (250 AlpacaEval \cite{li2023alpacaeval}, 100 Arena-Hard-Auto \cite{li2024crowdsourced}, 304 YapBench \cite{borisov2026yapbench} and the 16 MT-Bench development prompts), and tested on two fresh frames drawn after v2 was fixed: 250 prompts, and a further 705 prompts never generated for before. Answers are capped at 768 tokens (1,024 for the processor).

\paragraph{Measures} The token reduction ratio is the ratio of sums $\TRR=1-\sum_i n_i^{\mathrm{method}}/\sum_i n_i^{\mathrm{baseline}}$, counting generated tokens for the processor and shown tokens for \elide{}. The processor is tested for BERTScore \cite{zhang2020bertscore} non-inferiority against pre-registered margins of 0.20, 0.18 and 0.15, with question-clustered bootstrap bounds. For \elide{}, the baseline answer's sentences are content units, and \emph{NLI coverage} is the mean over units of their best entailment by a sentence of the method's answer, scored by RoBERTa-large-MNLI \cite{liu2019roberta, williams2018broad}, a judge that shares no weights with the licence; a unit with entailment below 0.5 is \emph{lost}. Coverage is reported beside the judge's \emph{ceiling}, the baseline scored against itself. Three pre-registered gates must pass: \emph{information} (coverage $\ge 0.95$ or within 0.01 of the ceiling), \emph{facts and structure} (numbers, entities, lists, code and tables preserved) and \emph{truncation} (coverage above the matched-length truncation baseline, exact Wilcoxon signed-rank test). A pre-registered \emph{reading} labels each withheld span, in context, as restating kept text, a closing, a restated question, a loss of specificity, a loss of content or damage to form; the last two are \emph{harmful} (Wilson 95\% intervals).

\section{Results}
\label{sec:results}

\subsection{The processor clears its margins, and is truncation}
\label{sec:res-proc}

The processor meets its pre-registered claim (TRR $\ge$ 12\% and both BERTScore bounds above their margins) on all three models, on the development and the sealed test split (Table~\ref{tab:proc}), although the answers it changes lose 0.10--0.18 of BERTScore recall on average.

\begin{table}[!t]
\caption{The processor at its frozen operating point on MT-Bench. NI: one-sided 95\% lower bound of the method-minus-baseline BERTScore difference (turn 1) against the margin $-\delta$; all bounds pass. Unchanged: turn-1 test answers identical to the baseline.}
\label{tab:proc}
\begin{tabular*}{\hsize}{@{\extracolsep{\fill}}lccccc@{}}
\toprule
Model & TRR dev [95\% CI] & TRR test [95\% CI] & $\delta$ & NI recall, dev / test & Unchanged (test) \\
\midrule
Mistral-7B   & 25.5\% [14.2, 34.7] & 19.9\% [14.2, 25.1] & 0.20 & $-0.165$ / $-0.012$ & 35 of 64 \\
Llama-3.1-8B & 34.9\% [20.6, 43.7] & 34.2\% [28.7, 39.2] & 0.18 & $-0.103$ / $-0.000$ & 32 of 64 \\
Qwen2.5-3B   & 30.2\% [16.2, 40.2] & 32.6\% [27.5, 37.0] & 0.15 & $-0.097$ / $+0.000$ & 35 of 64 \\
\bottomrule
\end{tabular*}
\end{table}

\begin{figure}[!htb]
\centering
\includegraphics[width=0.62\textwidth]{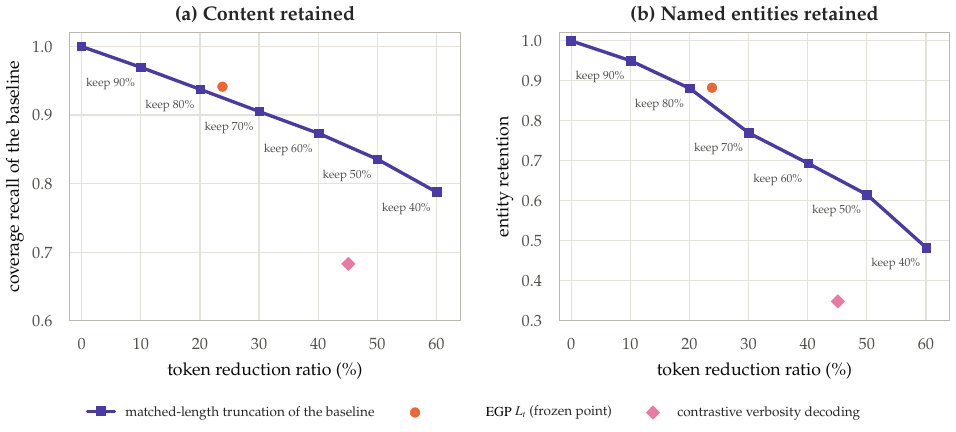}
\caption{The truncation frontier on Qwen2.5-3B (MT-Bench development): (a) coverage recall of the baseline and (b) entity retention against token reduction. The processor's length term sits on the frontier, as Proposition~\ref{prop:frozen} requires; contrastive verbosity decoding, which edits token choices, falls far below it.}
\label{fig:frontier}
\end{figure}

A decision-level audit (Mistral-7B, 17 prompts, 4,724 decoding steps) confirms Proposition~\ref{prop:frozen}: the processor changed 8 decisions, all 8 to a stop token, and its largest change to any non-stop logit was exactly 0.0. Restoring only the stop-token columns reduced the TRR from 0.260 to 0.000. Every changed answer equalled the baseline's first $k$ tokens, $k\in[385,400]$, followed by a stop, and seven of the eight cuts fell mid-sentence. None of the 12 redundant sentences in these answers was removed, while all 56 removed sentences were novel. On all 192 turn-1 answers of the sealed test split, the output is a token prefix of the baseline. The processor is an \emph{entropy-gated stop controller}: it ends answers at a configured position, not at a point of completeness, and its quality is by construction that of truncation (Fig.~\ref{fig:frontier}).

\subsection{No training-free mechanism beats truncation by selection}
\label{sec:res-survey}

We then tested 23 pre-registered families of training-free mechanisms against the frontier, including concise prompting, contrastive decoding and activation steering, minimum-Bayes-risk and best-of-$k$ selection, beam search and lookahead, post-hoc extraction, learned scoring over hidden states, and clause- and span-level deletion. Apart from stop controllers, every family tied truncation, \emph{drifted} (answered a smaller question), \emph{rerouted} (substituted different content; e.g.\ steering retained only 37.7--48\% of entities), lacked sample diversity, or found too little of its target. Two diagnostics explain the pattern: given an oracle utility signal, the model acts on it perfectly (roll-out AUROC 1.000), but asked which of its own sentences are redundant, it performs at chance (AUROC 0.495). The model can act on a redundancy signal but cannot compute one. Even a perfect sentence-level oracle beats truncation by only $+0.017$ to $+0.024$ BERTScore recall: the headroom for selection is small.

\subsection{\elide{} beats matched-length truncation}
\label{sec:res-elide}

On the 8 screen prompts (60 arms, Mistral-7B), cosine similarity alone removed 4.9\% of tokens but lost 8 content units by flagging template-sharing sentences; the licence at $\tau=0.75$ refused 20 such spans (entailment below 0.43), lost none, and was selected under the pre-registered rule. No combination with the length term passed the information gate. Table~\ref{tab:heldout} reports the held-out verification on three models.

\begin{table}[!htb]
\caption{Held-out prompts (24 per model). Cov.: NLI coverage; Ceiling: the baseline scored against itself. $\Delta$ trunc.: paired coverage gain over matched-length truncation on the answers each arm shortens, with wins--losses. Gates: information, facts and structure, truncation (y = pass); the length term fails all three.}
\label{tab:heldout}
\begin{tabular*}{\hsize}{@{\extracolsep{\fill}}lcccccccc@{}}
\toprule
 & Ceiling & \multicolumn{4}{c}{\elide{} v1} & \multicolumn{3}{c}{Length term (frozen point)} \\
Model & Cov. & TRR & Cov. & $\Delta$ trunc. (W--L) & Gates & TRR & Cov. & $\Delta$ trunc. \\
\midrule
Mistral-7B   & 0.958 & 0.9\% & 0.956 & $+0.034$ (4--0)  & yyy & 25.0\% & 0.785 & $0.000$ \\
Llama-3.1-8B & 0.988 & 2.2\% & 0.982 & $+0.029$ (9--0)  & yyy & 12.3\% & 0.886 & $0.000$ \\
Qwen2.5-3B   & 0.989 & 1.9\% & 0.982 & $+0.027$ (10--0) & yyy & 29.3\% & 0.753 & $0.000$ \\
\bottomrule
\end{tabular*}
\end{table}

\elide{} passes all three gates on every model, removing 0.9--2.2\% of shown tokens (pooled 1.71\%, 95\% CI [1.08, 2.41]) at a coverage within 0.007 of the ceiling. It never loses to matched-length truncation on an answer it shortens (exact Wilcoxon $p=0.004$ and $0.002$ on Llama and Qwen) and beats length-only and random deletion on average, while the length term ties truncation exactly. All 40 held-out elisions were read: 29 repeat kept text (e.g.\ ``Use a comfortable mattress and pillow.'' after ``Use comfortable bedding and pillows.''), 6 are closings or restated questions, 4 lose a specific detail and 1 is arguable. The reduction is small because there is little to remove: a pinned duplicate detector marks only 1.9--4.8\% of held-out sentences as restatements.

\subsection{At scale, the automatic measures miss the harm}
\label{sec:res-scale}

On the 670-prompt development frame (2,010 answers), v1 again looked sound by every automatic measure: it removed 2.65--2.93\% of shown tokens, beat matched-length truncation on every model (Wilcoxon $p<10^{-10}$), and what it withheld was 2.9--3.6 times as likely to be a restatement as an average sentence. The pre-registered reading disagreed: of 936 elisions read, 59.4\% [56.2, 62.5] lost content or damaged form (49\%, 69\% and 60\% by model). The harm fell into four classes that the hand-written prompts had not contained: (1)~an item's own attribute repeated across list items, e.g.\ ``It is a unisex name.'' after each of several names, which the licence correctly judges entailed by an identical sentence about \emph{another} name; (2)~the answer itself withheld as a restatement of the question (``The chemical symbol for sodium is Na.''), 84\% of such elisions being harmful; (3)~repetition that is the deliverable, such as a chorus or speaker labels; and (4)~fragments cut inside tokens such as ``e.g.'' or ``.357''. A second, LLM-based recoverability judge (Phi-3.5-mini) passed its controls yet rated about nine in ten of the harmful elisions as recoverable, separating harmful from acceptable elisions barely above chance (AUROC 0.53--0.64).

\begin{figure}[!htb]
\centering
\includegraphics[width=0.5\textwidth]{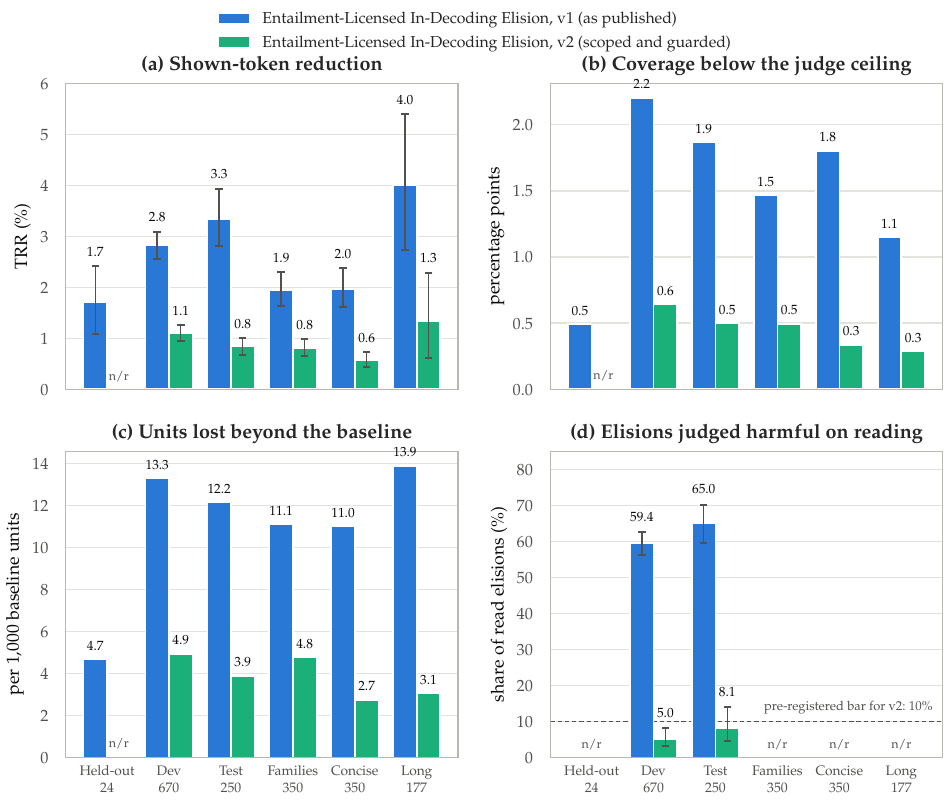}
\caption{\elide{} across evaluation sets, pooled over models: (a) shown-token reduction with 95\% intervals; (b) NLI coverage below the judge's ceiling; (c) baseline units lost beyond the baseline's own, per 1,000; (d) share of read elisions judged harmful, with Wilson intervals and the pre-registered 10\% bar for v2. Families: OLMo-2 and Granite-3.1; Concise: answers already prompted to be concise; Long: answers regenerated with a 2,048-token cap. n/r: not run or not read.}
\label{fig:main}
\end{figure}

Version 2 addresses one harm class per restriction. Replayed on the same 2,010 answers, it withholds 298 spans instead of 1,023, passes every gate on every model, and only 5.0\% [3.1, 8.1] of its elisions are harmful (Fig.~\ref{fig:main}). Because v2 was built by reading this frame, it was tested on two fresh frames under a pre-registered claim rule: ``v2 rarely harms'' would be licensed only if the Wilson upper bound of its harmful share were at most 10\%. On the first fresh test (250 prompts, 750 answers), v1's harm replicated (65.0\% of 300 read), while v2's harmful share fell to 8.1\% [4.6, 13.9]; on the second (705 prompts, 1,175 answers) it was 11.9\% [8.3, 16.8], with v2 again passing every gate and removing 0.96\% of shown tokens against v1's 3.41\%. Pooled over both, 37 of 363 v2 elisions are harmful, 10.2\% [7.5, 13.7]. By its own rule, the strong safety claim is therefore not licensed: v2 harms about one elision in ten.

\paragraph{The harm is a matter of scope, not of threshold} Raising a threshold is the obvious response. Replaying v1's 670-prompt decisions offline against the reading labels, raising the licence threshold $\theta$ from 0.5 to 0.99 cuts the reduction from 2.82\% to 0.67\% but leaves the harmful share of the surviving elisions at 59--61\%; raising the cosine threshold $\tau$ to 0.98 makes it \emph{worse}, 87.8\%, because the most similar spans are disproportionately verbatim attributes repeated across items. The licence is confident precisely where it is wrong, since two identical sentences do entail each other: the error lies in the premise it is given, which is why v2 restricts the licence's scope and leaves both thresholds unchanged.

\subsection{Transfer, licence reliability and cost}
\label{sec:res-transfer}

On OLMo-2 and Granite-3.1, never used in development, v2 removes 1.03\% and 0.55\% of shown tokens, passes every gate and beats truncation on 55 of 66 decided answers. On answers already prompted to be concise it removes a further 0.57\%, so prompting and elision are complementary, and its reduction grows with answer length (1.34\% at a 2,048-token cap, where 12.5\% of sentences are restatements). Under sampling, batching, four precisions, the published answers of six closed systems and machine-translated prompts in six languages including Arabic, v2's automatic behaviour transfers (0.56--2.90\% TRR per condition), but a sampled reading finds a higher harmful share in every family (9.7--19.4\%). On a contrast set built from its own decisions, the safeguards catch nearly all number, entity and added-content edits, but 13--19\% of negation, antonym and quantifier edits and 49\% of modal edits would still be withheld; yet a perfect licence would prevent only about a quarter of the residual harm. The cost is small: 2.4--2.9\% of generation time at batch size one, and 1.0--2.6\% at batch sizes 8--32 when the encoder and licence are batched.

\section{Discussion and limitations}
\label{sec:discussion}

The central lesson is methodological: a verbosity controller that passes conventional quality margins can be nothing more than a learned stopping position, which reference-anchored similarity rewards by construction, so every decode-time compression method should be compared with matched-length truncation. Beating the frontier is necessary but not sufficient: \elide{} v1 beat truncation on hundreds of prompts while most of its elisions harmed the answer, and two automatic judges of different kinds missed the harm. Safety claims for omission methods must rest on reading, not on automatic judges.

Positively, restatement can be withheld during decoding with an inspectable licence and without changing generation, and scoping the licence to its discourse unit, as ellipsis requires, removes most of the harm. The saving is small because ordinary answers contain little restatement (2--6\% of sentences). On two small base models the excess was loops rather than restatement; the length term removed 68--77\% of a looping model's tokens without loss, yet still tied truncation.

\emph{Limitations.} Main results use greedy decoding at batch size one; the transfer study relaxes each choice but reads a sample. Readings were done within the study, not by external annotators, and the multilingual prompts are machine-translated. \elide{} saves shown, not generated, tokens: judging a sentence from its first $k$ tokens reached a precision of at most 0.43 against the full-sentence decision.

\section{Conclusion}
\label{sec:conclusion}

Measured against matched-length truncation, most training-free verbosity reduction in LLM decoding is truncation in disguise. An entropy-gated processor that removes a quarter to a third of tokens within pre-registered margins provably only chooses where to stop, and 23 further mechanism families do no better at selecting content. Entailment-licensed elision is the one method that beats the frontier, by withholding spans whose content the reader has already seen; at scale most of its first version's elisions were harmful, and a discourse-scoped licence reduces this to about one elision in ten. Future work should derive the licence's scope from discourse structure and evaluate elision on long-form and retrieval-augmented generation with external readers.

\bibliographystyle{elsarticle-num}
\bibliography{references}

\end{document}